\documentclass{article}

\usepackage{microtype}
\usepackage{graphicx}
\usepackage{subcaption}

\usepackage[utf8]{inputenc} 
\usepackage[T1]{fontenc}    
\usepackage{hyperref}       
\usepackage{url}            
\usepackage{booktabs}       
\usepackage{amsfonts}       
\usepackage{nicefrac}       
\usepackage{microtype}      
\usepackage{xcolor}         

\usepackage{amsmath}
\usepackage{amssymb}
\usepackage{mathtools}
\usepackage{amsthm}
\usepackage[preprint]{neurips_2026}
\title{MindAU: EEG-Conditioned Facial Action Unit Editing via Dual-Stream Manifold Alignment}

\author{%
  Zhenhang Li$^{\dagger}$ \\
  Binghamton University \\
  \texttt{zli74@binghamton.edu} \\
  \And
  Xin Zhou$^{\dagger}$ \\
  Binghamton University \\
  Massachusetts General Hospital \\
  \texttt{xzhou11@binghamton.edu} \\
  \And
  Hao Deng$^{*}$ \\
  Massachusetts General Hospital \\
  Harvard Medical School \\
  \texttt{hdeng1@mgh.harvard.edu} \\
  \And
  Lijun Yin$^{*}$ \\
  Binghamton University \\
  \texttt{lijun@cs.binghamton.edu} \\
}

\begin{document}

\maketitle

\begingroup
\renewcommand\thefootnote{}\footnotetext{$^{\dagger}$ Equal contribution. $^{*}$ Corresponding authors.}
\endgroup

\begin{abstract}

Recent brain decoding studies have made substantial progress in reconstructing externally perceived visual content from neural signals. However, using electroencephalography (EEG) recordings to guide facial expression editing remains largely unexplored and poses a distinct challenge: rather than recovering what a subject sees, it requires identifying facial-action related patterns from noisy EEG signals and grounding them in localized, identity-preserving expression edits. In this paper, we investigate EEG-conditioned facial image editing for fine-grained facial action unit (AU) control and propose \textbf{MindAU}, a unified framework for controlling facial AU edits from EEG signals. MindAU first learns noise-robust and AU-discriminative EEG representations through temporal masked reconstruction and AU classification supervision. It then bridges the modality gap via Dual-Stream Manifold Alignment, aligning EEG features with AU-level text semantics and identity-reduced visual displacement trajectories in the multimodal space of Qwen2.5-VL. Finally, MindAU incorporates EEG-aware Multimodal Rotary Positional Embeddings, landmark-guided reference masking, and AU-aware region supervision into a multimodal diffusion-based editor for high-fidelity identity-preserving editing. We also introduce \textbf{E-CAFE}, a curated benchmark for EEG-Conditioned Action-Unit Facial Editing with paired EEG-face editing samples and standardized evaluation protocols. Extensive experiments demonstrate the effectiveness of MindAU and suggest its potential as a step towards future assistive expression technologies for individuals with facial neuromuscular disorders.

\end{abstract}

\section{Introduction}

Recent brain decoding studies have shown remarkable progress in reconstructing externally perceived visual content from neural signals, ranging from static images~\cite{mindaligner, mindbridge, kong2024toward} to dynamic videos~\cite{mind-animator, mind-video, eeg2video}. Most existing methods focus on recovering what a subject sees, typically by aligning neural representations with visual or semantic spaces through contrastive learning~\cite{sun2023contrast, mind-eye} or signal reconstruction~\cite{dreamdiffusion}. However, an underexplored and potentially impactful direction is to move beyond passive perceptual reconstruction towards controllable facial behavior editing conditioned on brain activity, establishing a pathway from neural signals to interpretable facial actions. This is especially relevant when paired neural recordings and facial-expression observations are scarce: electroencephalography (EEG)-conditioned facial editing may enrich EEG--face supervision and facilitate cross-modal modeling between brain activity and facial dynamics. It may also support future internal-state modeling, such as affect- or pain-related analysis, and assistive communication interfaces where expressive facial behavior is driven by neural signals.

Facial expressions can be decomposed into facial action units (AUs), providing a structured and interpretable control space for linking neural signals to localized facial movements. Unlike reconstructing perceived stimuli, EEG-conditioned facial editing requires extracting subtle and noisy facial-action cues from EEG signals and grounding them in fine-grained, identity-preserving expression changes on a reference face, posing a distinct and challenging problem. To the best of our knowledge, no prior work has addressed EEG-conditioned facial image editing with AU-level control. To fill this gap, we propose MindAU, the first framework for EEG-conditioned facial image editing at the action-unit level, which also serves as an early step towards future assistive expression technologies for individuals with facial neuromuscular disorders~\cite{Bell, social, impact}.

Despite its promise, EEG-driven facial expression editing is challenged by noisy signals, cross-modal alignment gaps, identity-preserving editing, and limited paired EEG-face data. To address these challenges, we propose MindAU, an EEG-conditioned generative framework for identity-preserving facial action unit editing. \textbf{First}, an AU-Aware EEG Encoder learns noise-robust and AU-discriminative representations via temporal masked reconstruction and AU classification supervision. \textbf{Second}, a Dual-Stream Manifold Alignment module maps EEG features to Qwen2.5-VL's multimodal space, aligning them with both AU-level text semantics and identity-reduced visual displacement trajectories. \textbf{Third}, we incorporate EEG-aware M-RoPE, landmark-guided reference masking, and AU-aware region supervision into a multimodal diffusion-based editor to enable EEG-driven local expression editing while preserving reference identity. \textbf{Finally}, we introduce \textbf{E-CAFE}, a curated benchmark built upon BU-EEG~\cite{BUEEG}, containing 5,000 paired EEG--face editing samples and two standardized protocols for self-referenced and cross-identity evaluation.

The primary contributions are summarized as follows:
\begin{itemize}
    \item We formulate EEG-conditioned facial expression editing as a fine-grained, identity-preserving facial action unit editing task. To the best of our knowledge, MindAU is the first framework specifically designed for EEG-conditioned action-unit-level facial image editing.
    \item We introduce an AU-Aware EEG Encoder with temporal masked reconstruction and AU classification supervision for noise-robust, AU-discriminative EEG representation learning.
    \item We propose Dual-Stream Manifold Alignment to bridge EEG representations with Qwen2.5-VL's multimodal space through AU-level text semantics and identity-reduced visual displacement trajectories.
    \item We adapt a multimodal diffusion-based editing backbone with EEG-aware M-RoPE, landmark-guided reference masking, and AU-aware region supervision to mitigate shortcut learning and preserve reference identity.
    \item We construct \textbf{E-CAFE}, a curated benchmark built upon BU-EEG with 5,000 paired EEG--face editing samples and two standardized evaluation protocols.
\end{itemize}

\section{Related Work}

\subsection{Representation Learning for EEG Signals}

Learning robust representations from noisy EEG signals is essential for brain decoding. Existing pre-training strategies mainly fall into two categories. Generative methods model EEG signal distributions through raw waveform reconstruction~\cite{dreamdiffusion, neurorvq} or time-frequency codebook learning~\cite{neuroLM, labram}. Task-oriented approaches, often based on contrastive learning~\cite{objection, braindreamer, eeg-clip}, map EEG representations into downstream spaces such as semantic classification~\cite{brainvis} or cross-modal retrieval~\cite{ATM}. However, these methods are not explicitly grounded in facial action semantics. EEG-conditioned facial editing requires noise-robust and AU-discriminative features to control local facial movements from neural signals. We therefore introduce AU-aware EEG pre-training with temporal masked reconstruction and AU classification supervision.

\begin{figure*}[ht]
  \centering\includegraphics[width=1\linewidth, trim=40 30 40 20, clip]{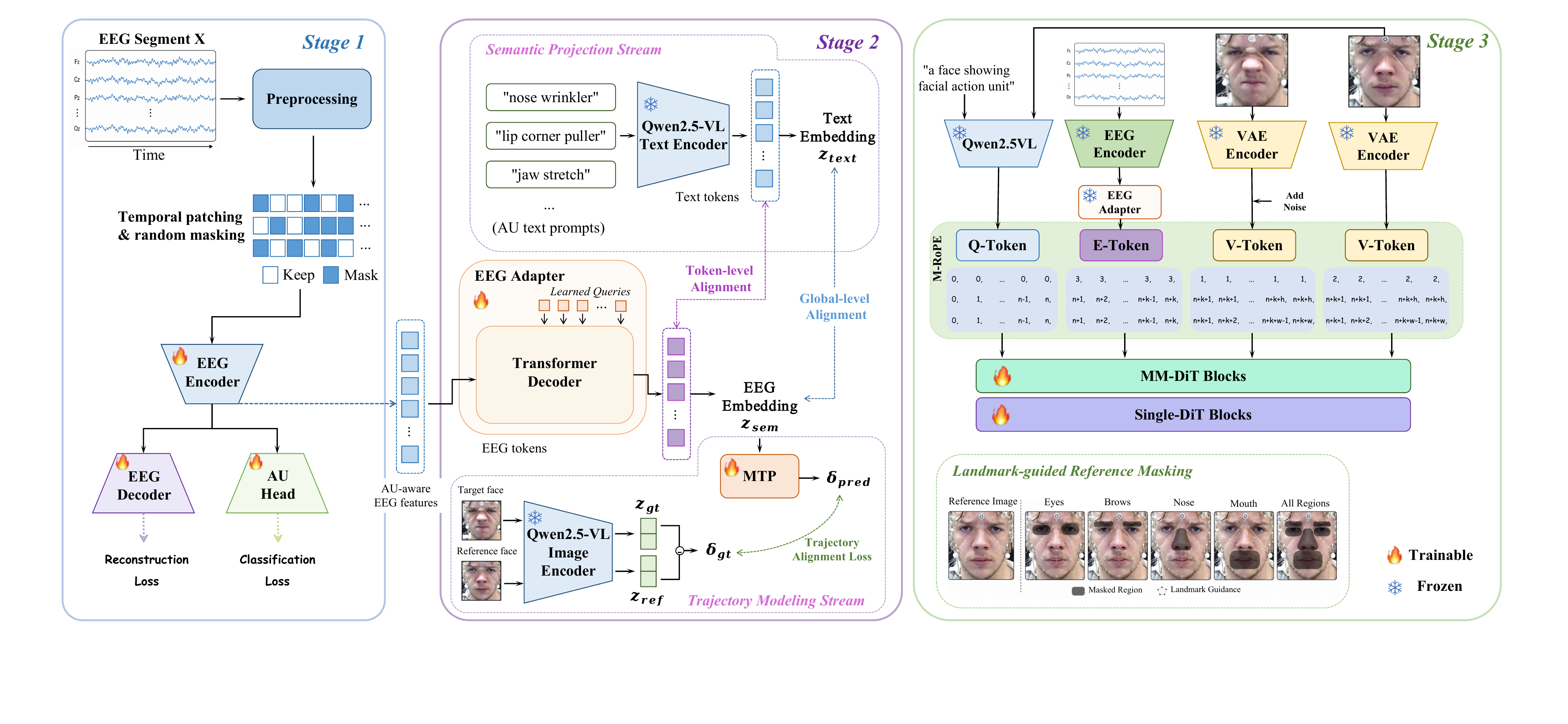}
  \vspace{-30pt}
  \caption{Overall architecture of MindAU. MindAU learns AU-discriminative EEG representations, aligns them with AU-level semantics and identity-reduced visual displacement trajectories, and conditions a multimodal diffusion editor for identity-preserving facial AU editing. 
MTP denotes the Manifold Trajectory Projector.}
  \vspace{-10pt}
  \label{fig:pipeline}
\end{figure*}

\subsection{Brain-Guided Image Generation and Editing}

Brain-guided visual generation synthesizes images from neural signals such as fMRI and EEG. Early methods used GANs~\cite{brain2image} to reconstruct perceived stimuli, whereas recent approaches adopt diffusion models~\cite{zeng2024controllable, dreamdiffusion, brainvis}. Most existing works focus on passive perceptual reconstruction from fMRI~\cite{mind-vis, mind-eye} or EEG~\cite{braindreamer} recorded during stimulus viewing. In parallel, text-guided diffusion models~\cite{flux, sd3} and facial editing models~\cite{longcat-image, zimage,lizhen} have achieved high-fidelity semantic editing with strong identity preservation, but they rely on explicit text or image conditions and cannot be directly driven by noisy brain signals. Although several studies use brain signals as implicit guidance for image editing~\cite{davis2022brain, mindpainter, mind-artist, mindcustomer}, they mainly rely on neural responses to external stimuli rather than grounding facial-action-related EEG cues in localized expression edits. Recent work has also explored EEG-based facial synthesis. Mind-to-Face~\cite{mind} maps EEG signals to dense 3D facial geometry for subject-specific avatar synthesis, whereas our method targets reference-based facial image editing with transferable AU-level control and identity preservation.

\section{Methodology}


Given an EEG segment $\mathbf{X}$ and a reference face image $\mathbf{I}_{ref}$, MindAU generates an edited image $\mathbf{\hat{I}}$ that preserves the identity in $\mathbf{I}_{ref}$ while expressing the facial action implied by $\mathbf{X}$. It proceeds in three stages: (i) \textit{Action-Unit-Aware EEG Pre-training} learns AU-discriminative EEG representations; (ii) \textit{Dual-Stream Manifold Alignment} maps these representations to semantic embeddings and identity-reduced visual displacement trajectories; and (iii) \textit{EEG-Conditioned Facial Editing}, which conditions an MMDiT-based face editing backbone on the aligned EEG tokens and the reference face to produce identity-preserving expression edits. We also introduce \textbf{E-CAFE}, a benchmark for \textbf{E}EG-\textbf{C}onditioned \textbf{A}ction-unit \textbf{F}acial \textbf{E}diting.

\subsection{AU-Aware EEG Encoder Pre-training}

\textbf{EEG preprocessing.}
For each trial, we remove DC offset, apply a fourth-order Butterworth band-pass filter between 0.1 and 50 Hz, and perform channel-wise Z-score normalization. This standard preprocessing removes slow drift, attenuates high-frequency noise, and normalizes channel-wise signal scales before temporal masked modeling.

\textbf{Temporal Masked Modeling.}
To learn robust representations from noisy EEG signals, we adopt a temporal Masked Autoencoder (MAE) backbone~\cite{mae}, following DreamDiffusion~\cite{dreamdiffusion}. Let $\mathbf{X} \in \mathbb{R}^{C \times T}$ denote a preprocessed EEG segment with $C$ channels and $T$ time points. We partition $\mathbf{X}$ into temporal patches and randomly mask a subset to obtain $\tilde{\mathbf{X}}$. The encoder $E_\theta$ maps $\tilde{\mathbf{X}}$ to $\mathbf{z}=E_\theta(\tilde{\mathbf{X}})$, and the decoder $D_\phi$ reconstructs the masked patches from $\mathbf{z}$. This objective encourages contextual temporal modeling rather than reliance on local noisy fluctuations:
\begin{equation}
    \mathcal{L}_{\mathrm{rec}} = \|\mathbf{M} \odot (\mathbf{X} - D_\phi(\mathbf{z}))\|_2^2,
\end{equation}
where $\mathbf{M}$ is the temporal patch mask and $\odot$ denotes element-wise multiplication.


\textbf{AU Semantic Supervision.}
Since signal reconstruction lacks explicit AU-level semantic grounding, we add an AU classification objective to learn AU-discriminative EEG features. Specifically, AU prediction is formulated as a multi-class classification task over predefined AU categories. The encoder feature $\mathbf{z}$ is fed into a classification head $f_{cls}(\cdot)$, and the loss is
\begin{equation}
    \mathcal{L}_{cls} = \mathrm{CE}(f_{cls}(\mathbf{z}), y),
\end{equation}
where $y$ is the ground-truth AU label and $\mathrm{CE}(\cdot,\cdot)$ denotes cross-entropy. The Stage-1 objective is
\begin{equation}
    \mathcal{L}_{stage1} = \mathcal{L}_{rec} + \lambda_{cls}\mathcal{L}_{cls},
\end{equation}
where $\lambda_{cls}$ balances reconstruction and AU supervision. This pre-training yields temporally contextualized and AU-aware EEG representations for subsequent cross-modal alignment.

\subsection{Dual-Stream Manifold Alignment}

Although the Stage-1 encoder captures AU-correlated EEG features, its latent space is not directly compatible with the semantic space used by pretrained multimodal generative models. We therefore introduce \textit{Dual-Stream Manifold Alignment} to bridge this gap. The \textit{Semantic Projection Stream} maps EEG features to the text-semantic space of Qwen2.5-VL, while the \textit{Trajectory Modeling Stream} aligns them with identity-reduced visual displacement trajectories for fine-grained editing.

\textbf{Semantic Projection Stream.}
We introduce an EEG Adapter to convert Stage-1 EEG features into semantic tokens compatible with Qwen2.5-VL~\cite{qwen2.5vl}. Given the EEG feature sequence, $Q=32$ learnable query tokens cross-attend to it through a Transformer Decoder. These queries are randomly initialized and optimized with the EEG Adapter. The resulting query features are projected into the embedding space of the frozen Qwen2.5-VL text encoder, forming the EEG semantic tokens. We use their pooled representation $\mathbf{z}_{sem}$ for global-level alignment and retain the full token sequence as E-tokens for token-level supervision and Stage-3 editing.

As semantic targets, we encode AU-specific natural language descriptions with the frozen Qwen2.5-VL text encoder and denote the pooled text embedding as $\mathbf{z}_{text}$. We first align $\mathbf{z}_{sem}$ and $\mathbf{z}_{text}$ using a cross-modal supervised contrastive loss:
\begin{equation}
\begin{split}
\mathcal{L}_{align} =
-\frac{1}{B}\sum_{i=1}^{B}\frac{1}{|P(i)|}\sum_{j\in P(i)}
\log
\frac{\exp(\mathrm{sim}(\mathbf{z}_{sem,i}, \mathbf{z}_{text,j})/\tau)}
{\sum_{k=1}^{B}\exp(\mathrm{sim}(\mathbf{z}_{sem,i}, \mathbf{z}_{text,k})/\tau)},
\end{split}
\end{equation}
where $P(i)$ denotes the set of text samples sharing the same AU label as EEG sample $i$, $\mathrm{sim}(\cdot,\cdot)$ is cosine similarity, and $\tau$ is the temperature.

To preserve token-level semantic structure, we further align EEG tokens with teacher text tokens. Let $\mathbf{E}_i=\{\mathbf{e}_{i,1},\dots,\mathbf{e}_{i,Q}\}$ be the EEG tokens of sample $i$, and $\mathbf{T}_i=\{\mathbf{t}_{i,1},\dots,\mathbf{t}_{i,L}\}$ be its valid teacher text tokens. For each EEG token, we keep its maximum similarity to any valid text token:
\begin{equation}
s_{i,q}=\max_{l:\,\mu_{i,l}=1}\mathrm{sim}(\mathbf{e}_{i,q},\mathbf{t}_{i,l}),
\end{equation}
where $\mu_{i,l}$ is the text validity mask. The token-level alignment loss is
\begin{equation}
\mathcal{L}_{tok}=1-\frac{1}{BQ}\sum_{i=1}^{B}\sum_{q=1}^{Q} s_{i,q},
\end{equation}
where $Q$ is the number of EEG tokens. This loss avoids enforcing one-to-one positional correspondence and instead allows each EEG token to align with its most relevant text token, preserving local semantic structure beyond the pooled embedding.

\textbf{Trajectory Modeling Stream.}
Although the Semantic Projection Stream aligns EEG features with AU-level text semantics, text supervision is categorical and cannot capture within-AU variation, such as expression intensity. We therefore introduce a Trajectory Modeling Stream to supervise EEG features with identity-reduced visual displacement trajectories.

Given the visual embeddings of the reference image $\mathbf{z}_{ref}$ and ground-truth target image $\mathbf{z}_{gt}$ extracted by the frozen Qwen2.5-VL visual encoder~\cite{qwen2.5vl}, we compute the raw visual displacement as
\[
\boldsymbol{\delta}_{raw} = \mathbf{z}_{gt} - \mathbf{z}_{ref}.
\]
Since this displacement may contain identity-related components, we reduce the component aligned with the reference embedding by projecting it onto the orthogonal complement of $\mathbf{z}_{ref}$:
\begin{equation}
    \boldsymbol{\delta}_{gt}  = \boldsymbol{\delta}_{raw} - \frac{\boldsymbol{\delta}_{raw} \cdot \mathbf{z}_{ref}}{\|\mathbf{z}_{ref}\|^2} \mathbf{z}_{ref}.
\end{equation}
The resulting $\boldsymbol{\delta}_{gt}$ serves as an identity-reduced visual displacement target that emphasizes expression-specific deformation relative to the reference face.

We train the Manifold Trajectory Projector (MTP) to predict an EEG-driven trajectory $\boldsymbol{\delta}_{pred}$ from the pooled EEG representation. The trajectory loss combines directional consistency and magnitude matching:
\begin{equation}
    \mathcal{L}_{traj} = \lambda_{cos} \bigl(1 - \cos(\boldsymbol{\delta}_{pred}, \boldsymbol{\delta}_{gt})\bigr) + \frac{\lambda_{L1}}{D} \|\boldsymbol{\delta}_{pred} - \boldsymbol{\delta}_{gt}\|_1,
\end{equation}
where $D$ is the feature dimension, and $\lambda_{cos}$ and $\lambda_{L1}$ balance the two terms.

The overall Stage-2 objective is
\begin{equation}
    \mathcal{L}_{stage2}
    =
    \mathcal{L}_{align}
    +
    \lambda_{tok}\mathcal{L}_{tok}
    +
    \lambda_{traj}\mathcal{L}_{traj},
\end{equation}
where $\lambda_{tok}$ and $\lambda_{traj}$ balance token-level semantic alignment and trajectory supervision. To provide coarse magnitude-aware supervision for Stage-2 alignment, we synthesize identity-diversified expression-transition pairs with Nano Banana 2~\cite{Banana}. For each real sample, we use the reference-to-target facial change, including the target AU and observed expression strength, to guide the same intended transition on four generated identities. Rather than serving as exact trajectory targets, these pairs provide an approximate prior on EEG-associated deformation magnitude across identities, helping the EEG Adapter learn identity-invariant AU deformation and coarse intensity variation before fine-tuning on real EEG--face pairs.

\subsection{EEG-Conditioned Facial Editing}

With the aligned EEG representations, MindAU builds on a Multimodal Diffusion Transformer (MM-DiT)\cite{sd3} for identity-preserving facial editing. As shown in Figure~\ref{fig:pipeline} (Stage 3), a fixed text prompt and a reference image are encoded by the frozen Qwen2.5-VL to produce semantic Q-tokens, while the frozen EEG encoder and EEG Adapter map the EEG signal into 32 E-tokens. The ground-truth image is encoded by the VAE and perturbed with Gaussian noise to form the noisy latent V-token$_{noise}$, and the reference image is encoded as V-token$_{ref}$ to provide identity information.

To fuse these heterogeneous tokens in a shared attention space, we redesign Multimodal Rotary Positional Embedding (M-RoPE)\cite{longcat-image}. The positional encoding contains one modality dimension, which distinguishes text, EEG, ground-truth image, and reference image, and two spatial dimensions, which encode image geometry. Image tokens use their natural $(h,w)$ coordinates, while each 1D text or EEG token uses its sequence index for both spatial dimensions. Following LongCat-Image \cite{longcat-image}, the backbone consists of double-stream blocks followed by single-stream blocks.

We train the model with Conditional Flow Matching (CFM). Let $\mathbf{z}_0$ denote the latent of the ground-truth image and $\mathbf{z}_1 \sim \mathcal{N}(\mathbf{0},\mathbf{I})$ denote Gaussian noise. The interpolation path is
\begin{equation}
    \mathbf{z}_t = (1-t)\mathbf{z}_0 + t\mathbf{z}_1,
\end{equation}
with target velocity
\begin{equation}
    \mathbf{v}_{target} = \frac{\mathrm{d}\mathbf{z}_t}{\mathrm{d}t} = \mathbf{z}_1 - \mathbf{z}_0.
\end{equation}
Given the multimodal condition set
\begin{equation}
    \mathcal{C} = \{\text{Q-token}, \text{E-token}, \text{V-token}_{ref}\},
\end{equation}
the model $v_\theta$ predicts the velocity from $\mathbf{z}_t$, yielding the flow-matching loss
\begin{equation}
\mathcal{L}_{FM}
=
\mathbb{E}_{t,\mathbf{z}_0,\mathbf{z}_1}
\left[
\left\|
v_\theta(\mathbf{z}_t,t,\mathcal{C})
-
(\mathbf{z}_1-\mathbf{z}_0)
\right\|_2^2
\right].
\end{equation}
At inference, the model takes only the EEG segment, the reference image, and the fixed text prompt as conditions; the ground-truth image is used only for training.

Training on limited data can lead to a degenerate shortcut: as the reference image already accounts for most target-image content, the model may learn a reference-dominant near-identity mapping and under-utilize EEG conditions for expression control. To mitigate this issue, we introduce landmark-guided reference masking and AU-aware region supervision.

\textbf{Landmark-Guided Reference Masking.}
To reduce expression leakage from the reference branch, we mask AU-relevant regions in the reference image before encoding it as a visual condition. The masks are derived from reference-image landmarks and expanded to cover surrounding AU-related facial regions, including eyes, brows, nose, and mouth. We use expanded ellipses for eyes, brows, and mouth, and a convex hull for the nose; all masks are further dilated to cover nearby expression-related texture changes. We use a progressive masking schedule, decreasing the fraction of training samples with masked references from 70\% to 30\%. This encourages early reliance on EEG conditions while gradually restoring full reference information for identity-preserving editing at inference.

\textbf{AU-aware region supervision.}
In addition to the full-image flow-matching loss, we introduce two spatial losses computed from the same prediction residual. For the $b$-th sample, the channel-averaged spatial error map in latent space is
\begin{equation}
E_b(h,w)=\frac{1}{D}\sum_{c=1}^{D}
\left(
v_\theta(\mathbf{z}_{t,b},t,\mathcal{C}_b)_{c,h,w}
-
\mathbf{v}_{target,b,c,h,w}
\right)^2,
\end{equation}
where $D$ is the number of latent channels.

We first construct a facial-region mask by taking the union of the four AU-relevant regions:
\begin{equation}
M_{\mathrm{region}}^{(b)}=
\max\!\left(
M_{\mathrm{eye}}^{(b)},
M_{\mathrm{brow}}^{(b)},
M_{\mathrm{nose}}^{(b)},
M_{\mathrm{mouth}}^{(b)}
\right).
\end{equation}
The region loss averages prediction error over this facial action-relevant area:
\begin{equation}
\mathcal{L}_{\mathrm{region}}
=
\frac{1}{B}\sum_{b=1}^{B}
\frac{
\sum_{h,w}M_{\mathrm{region}}^{(b)}(h,w)\,E_b(h,w)
}{
\sum_{h,w}M_{\mathrm{region}}^{(b)}(h,w)+\epsilon
},
\end{equation}
where $\epsilon$ is a small constant for numerical stability.

We further define an AU-focused mask according to the target AU label:
\begin{equation}
M_{\mathrm{focus}}^{(b)}=
\begin{cases}
M_{\mathrm{brow}}^{(b)}, & a_b \in \{AU1, AU2, AU4\},\\
M_{\mathrm{eye}}^{(b)}, & a_b = AU5,\\
M_{\mathrm{nose}}^{(b)}, & a_b = AU9,\\
M_{\mathrm{mouth}}^{(b)}, & a_b \in \{AU12, AU15, AU17, AU25, AU27\}.
\end{cases}
\end{equation}
The AU-focused loss is
\begin{equation}
\mathcal{L}_{\mathrm{focus}}
=
\frac{1}{B}\sum_{b=1}^{B}
\frac{
\sum_{h,w}M_{\mathrm{focus}}^{(b)}(h,w)\,E_b(h,w)
}{
\sum_{h,w}M_{\mathrm{focus}}^{(b)}(h,w)+\epsilon
}.
\end{equation}

The final Stage-3 objective is
\begin{equation}
\mathcal{L}_{\mathrm{stage3}}
=
\mathcal{L}_{FM}
+
\lambda_{\mathrm{region}}\mathcal{L}_{\mathrm{region}}
+
\lambda_{\mathrm{focus}}\mathcal{L}_{\mathrm{focus}},
\end{equation}
where $\lambda_{\mathrm{region}}$ and $\lambda_{\mathrm{focus}}$ balance the two auxiliary losses.

\begin{figure}[ht]
  \centering\includegraphics[width=0.8\linewidth]{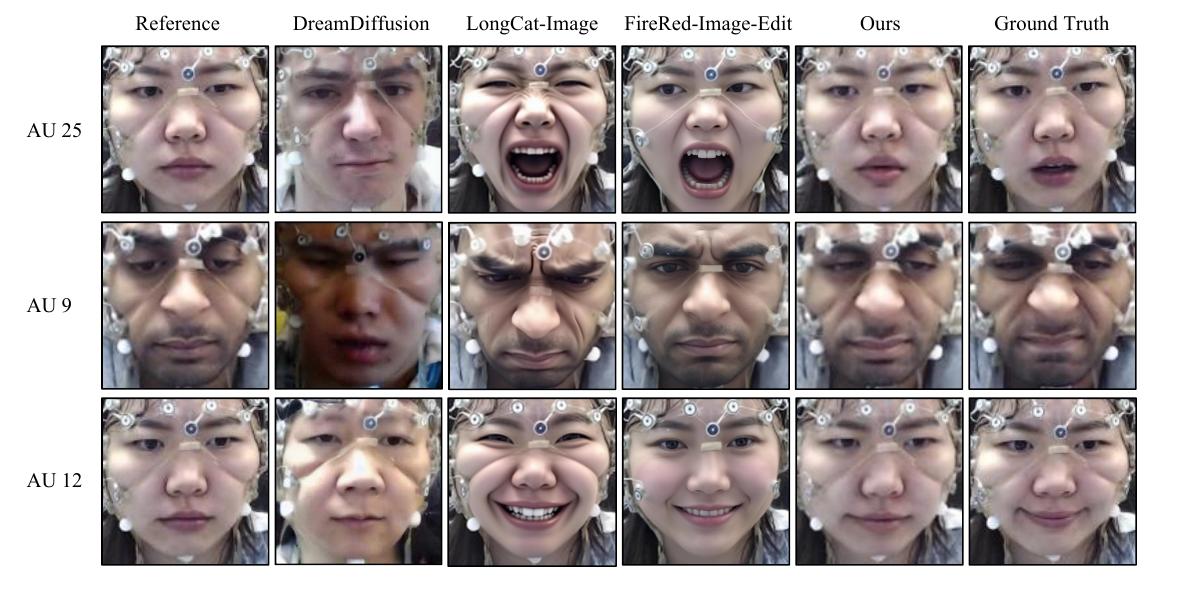}
  \vspace{-10pt}
  \caption{Visual comparison of our approach with other methods. The first, second, and third rows correspond to AU25 (Lips Part), AU9 (Nose Wrinkler), and AU12 (Lip Corner Puller).}
    \vspace{-5pt}
  \label{fig:comparison}
\end{figure}

\begin{figure}[ht]
  \centering\includegraphics[width=0.85\linewidth]{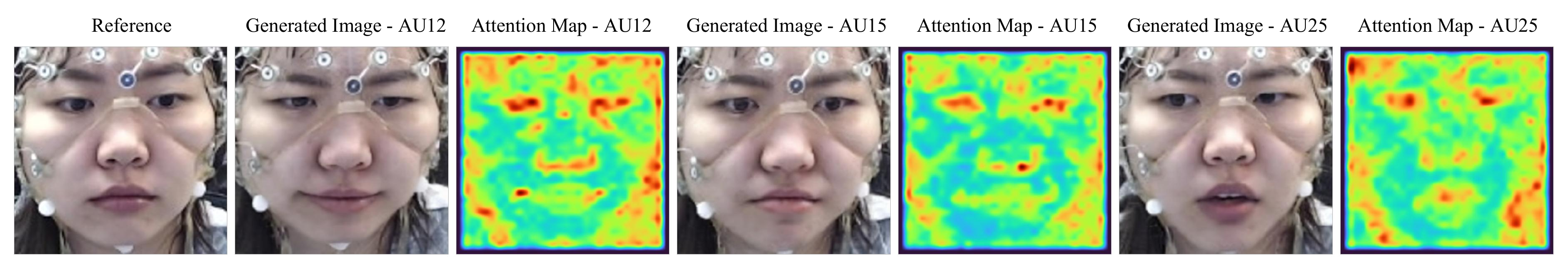}
  \vspace{-5pt}
  \caption{Visualization of EEG Attention–Sensitive Spatial Regions in Generated Images}
  \vspace{-15pt}
  \label{fig:attention}
\end{figure}
\subsection{E-CAFE: A Benchmark for EEG-Conditioned Action Unit Editing}
\label{sec:3.4}

To support standardized evaluation of EEG-driven facial editing, we introduce \textbf{E-CAFE} (\textbf{E}EG-\textbf{C}onditioned \textbf{A}ction-unit \textbf{F}acial \textbf{E}diting), a curated benchmark built upon BU-EEG~\cite{BUEEG}. E-CAFE provides paired EEG--face editing samples together with two evaluation protocols.

For subject-independent evaluation, we split BU-EEG by identity, reserving two subjects as a strictly held-out test set and using the remaining subjects for training. We first apply the Face Alignment Network (FAN)~\cite{fan} to detect facial landmarks and filter frames inconsistent with the target AU activation. For each retained target frame, we extract the corresponding preceding EEG segment of 48 time points and a reference face image. The resulting task is to transform the reference image into the target expression under the condition of the EEG segment. For testing, we further manually curate neutral-to-target sequences to reduce expression leakage from the reference image, ensuring that expression changes are primarily driven by EEG conditions.

E-CAFE includes two complementary evaluation protocols. \textbf{(1) Self-Referenced Editing} uses the reference image and EEG signal from the same held-out identity $(I_A, E_A)$. This setting evaluates reconstruction fidelity on unseen subjects by measuring whether the model can translate EEG signals into the corresponding facial dynamics while preserving identity. \textbf{(2) Cross-Referenced Editing} evaluates transferability and disentanglement. EEG signals from held-out subjects are used to drive 128 synthetic identities generated by Z-Image~\cite{zimage}, covering diverse ages and genders. This setting tests whether the learned EEG representations capture AU-related expression semantics independent of the source identity.

\section{Experiments}

\subsection{Datasets}
We utilize the BU-EEG dataset \cite{BUEEG}, which includes 28 participants performing 10 distinct AUs (AU1, AU2, AU4, AU5, AU9, AU12, AU15, AU17, AU25, AU27). It provides synchronized 128-channel EEG signals and facial videos sampled at 250 Hz and 24 fps, respectively.

\textbf{Evaluation Metrics.} We adopt a multi-dimensional protocol to comprehensively assess our framework. To quantify image quality and diversity, we compute the \textbf{Fréchet Inception Distance (FID)}~\cite{fid} to measure the distributional discrepancy between synthesized and ground-truth images. To evaluate identity preservation, we calculate the \textbf{Cosine Similarity (CSIM)} between the ArcFace~\cite{arcface} embeddings of the generated face and the reference identity. For expression fidelity, we report \textbf{AU Accuracy (AU ACC)} using OpenGraphAU~\cite{openau}, initialized from the model pre-trained on BP4D~\cite{bp4d} and further adapted on the E-CAFE training split. In addition, we report the \textbf{CLIP Score}~\cite{clipscore} to measure the semantic consistency between generated images and the corresponding conditioning descriptions. More details about the evaluation protocol are provided in the Appendix.


\subsection{Implementation Details}
Our training pipeline consists of three stages: (1) AU-Aware EEG Encoder Pre-training, (2) Dual-Stream Manifold Alignment, and (3) EEG-Conditioned Image Editing. All experiments are conducted on NVIDIA RTX A6000 GPUs. More details are provided in the Appendix.

\begin{figure}[ht]
  \centering\includegraphics[width=0.75\linewidth]{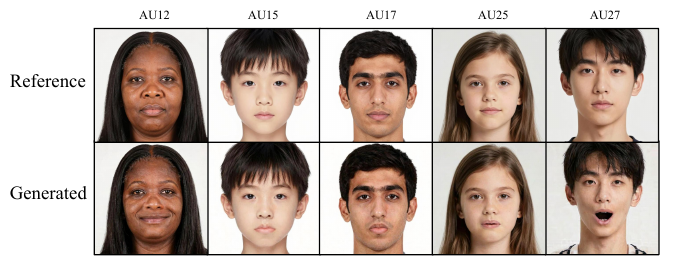}
  \vspace{-10pt}
  \caption{Qualitative results of Cross-Referenced Editing.}
  \vspace{-5pt}
  \label{fig:PART2}
\end{figure}

\subsection{Evaluation}




\begin{table*}[t]
\centering
\footnotesize
\setlength{\tabcolsep}{5pt}
\renewcommand{\arraystretch}{0.9}
\vspace{-5pt}
\caption{Quantitative comparison with existing methods. Pipeline baselines first predict AU categories from EEG and then edit images with fixed AU descriptions. ``GT Ref.'' denotes results on ground-truth target images.}
\label{tab:table1}
\begin{tabular}{lcccc}
\toprule
Method & CSIM(\%) $\uparrow$ & FID $\downarrow$ & AU ACC(\%) $\uparrow$ & CLIP Score $\uparrow$ \\
\midrule
DreamDiffusion~\cite{dreamdiffusion} & 16.59 & 106.25 & 9.23 & 13.85 \\
LongCat-Image~\cite{longcat-image} & 31.65 & 59.40 & 20.88 & \textbf{18.89} \\
FireRed-Image-Edit~\cite{firered} & 38.42 & 80.83 & 19.48 & 17.40 \\
Ours & \textbf{75.33} & \textbf{43.35} & \textbf{27.71} & 17.61 \\
\midrule
GT Ref. & \multicolumn{2}{c}{--} & 74.29 & 19.44 \\
\bottomrule
\end{tabular}
\vspace{-5pt}
\end{table*}

\begin{table*}[t]
\centering
\footnotesize
\setlength{\tabcolsep}{3.2pt}
\renewcommand{\arraystretch}{0.88}
\caption{Ablation studies under the Self-Referenced and Cross-Referenced Editing protocols. ``Direct'' and ``SupCon'' denote pairwise cosine EEG--text alignment and supervised contrastive global-level alignment, respectively.}
\label{tab:ablation}
\begin{tabular}{cccccc|cccc|ccc}
\toprule
\multicolumn{6}{c|}{Configuration} &
\multicolumn{4}{c|}{Self-Referenced} &
\multicolumn{3}{c}{Cross-Referenced} \\
\cmidrule(lr){1-6} \cmidrule(lr){7-10} \cmidrule(lr){11-13}
AU & Align & $\mathcal{L}_{tok}$ & $\mathcal{L}_{traj}$ & $\mathcal{L}_{reg}$ & $\mathcal{L}_{foc}$ &
CSIM$\uparrow$ & FID$\downarrow$ & AU ACC$\uparrow$ & CLIP$\uparrow$ &
CSIM$\uparrow$ & AU ACC$\uparrow$ & CLIP$\uparrow$ \\
\midrule
 & Direct &  &  &  &  &
74.96 & 51.09 & 13.25 & 14.92 &
57.88 & 9.74 & 19.37 \\

\checkmark & Direct &  &  &  &  &
71.81 & 45.85 & 15.66 & 16.24 &
54.70 & 11.14 & 20.27 \\

\checkmark & SupCon &  &  &  &  &
74.25 & 43.45 & 23.89 & 16.31 &
51.59 & 11.88 & 20.40 \\

\checkmark & SupCon & \checkmark &  &  &  &
74.70 & \textbf{41.87} & 20.88 & 16.02 &
52.89 & 11.86 & 20.43 \\

\checkmark & SupCon &  & \checkmark &  &  &
74.11 & 42.98 & 24.10 & 16.14 &
53.24 & 11.95 & 20.41 \\

\checkmark & SupCon & \checkmark & \checkmark &  &  &
73.50 & 43.59 & 24.71 & 16.26 &
53.48 & 12.01 & 20.52 \\

\checkmark & SupCon & \checkmark & \checkmark & \checkmark &  &
73.34 & 43.84 & 26.49 & 16.78 &
59.27 & 17.45 & 20.62 \\

\checkmark & SupCon & \checkmark & \checkmark & \checkmark & \checkmark &
\textbf{75.33} & 43.35 & \textbf{27.71} & \textbf{17.61} &
\textbf{66.21} & \textbf{20.70} & \textbf{21.09} \\
\bottomrule
\end{tabular}
\vspace{-2mm}
\end{table*}

\textbf{Quantitative Comparison.}
Since no existing method is designed for EEG-conditioned facial editing, we compare MindAU with three representative baselines under the Self-Referenced Editing protocol. DreamDiffusion~\cite{dreamdiffusion} is fine-tuned as an EEG-to-image baseline, generating images directly from test EEG signals without a reference image. We further construct two natural pipeline baselines with LongCat-Image~\cite{longcat-image} and FireRed-Image-Edit~\cite{firered}, decomposing EEG-conditioned editing into EEG-to-AU recognition followed by text-guided reference-based editing. For a fair comparison, the EEG-to-AU classifier used in these pipelines shares the same Stage-1 EEG encoder and AU classification head as MindAU, achieving 32.93\% standalone AU accuracy on the held-out test split. As shown in Table~\ref{tab:table1}, MindAU achieves the best performance on the three most task-critical metrics. Compared with DreamDiffusion, MindAU benefits from the reference image as an explicit identity condition and better preserves identity while translating EEG signals into facial actions. Compared with the two text-guided pipelines, MindAU avoids compressing EEG signals into a discrete AU label before editing. Fixed AU text prompts provide only coarse control and often lead to exaggerated facial deformations or unrecoverable errors when the EEG-to-AU classifier predicts an incorrect label. In contrast, MindAU conditions the editor on continuous EEG-aligned representations, enabling more AU-faithful, intensity-aware, and identity-preserving edits. LongCat-Image achieves the highest CLIP Score, likely due to its direct use of fixed AU text prompts. However, CLIP Score mainly measures coarse text-image consistency and does not fully reflect EEG faithfulness or expression-intensity alignment. We therefore interpret CLIP Score together with CSIM, FID, and AU ACC.

\textbf{Qualitative Comparison.}
Figure~\ref{fig:comparison} provides qualitative comparisons. DreamDiffusion often fails to preserve reference identity because it generates images directly from EEG without using the reference image. Text-guided pipelines better retain the reference structure but tend to produce canonical or overly strong expressions, such as exaggerated smiles for AU12, due to fixed textual AU prompts. They are also sensitive to AU recognition errors, where an incorrect predicted AU constrains the downstream editor. MindAU directly conditions generation on continuous EEG-aligned tokens. It produces subtler mouth movements for AU25 and AU12 and a nose-related deformation closer to the target for AU9, while better preserving identity. Figure~\ref{fig:attention} further shows EEG-sensitive attention maps concentrated on AU-relevant facial regions, indicating that EEG conditions guide local facial generation. Figure~\ref{fig:PART2} shows Cross-Referenced Editing results, where the same EEG signal drives consistent expression edits across different identities, suggesting that MindAU disentangles AU-related neural cues from facial identity. More results are provided in the Appendix.

\subsection{Ablation Studies}

We conduct ablations under both Self-Referenced and Cross-Referenced Editing protocols. As shown in Table~\ref{tab:ablation}, AU classification supervision improves AU ACC and CLIP Score, confirming that explicit AU labels help the EEG encoder learn expression-discriminative representations. Replacing direct pairwise alignment with supervised contrastive alignment brings a large AU ACC gain, highlighting the importance of AU-level semantic structure. 

$\mathcal{L}_{tok}$ and $\mathcal{L}_{traj}$ provide complementary supervision. The token-level loss preserves fine-grained semantic structure and yields the best FID, but its flexible token matching does not enforce global AU discrimination or deformation consistency when used alone. The trajectory loss introduces deformation-aware supervision, improving AU fidelity by aligning EEG features with identity-reduced visual changes. Stage-3 spatial losses further improve editing accuracy by concentrating optimization on AU-relevant regions. The full model achieves the best overall trade-off across CSIM, AU ACC, and CLIP Score. Table~\ref{tab:ablation} shows a similar trend under Cross-Referenced Editing. We further ablate landmark-guided reference masking in the Appendix.

\subsection{Limitations}
MindAU still has several limitations. First, although it has potential relevance to future assistive expression technologies, E-CAFE is built from healthy participants physically performing facial actions. These recordings may contain facial electromyographic artifacts that overlap with EEG frequency bands and are not fully removed by standard band-pass filtering. Thus, our results should be viewed as a first step towards EEG-conditioned facial AU editing rather than clinical validation on patients with facial paralysis. Future work should study imagined or attempted facial movements, patient populations, and stronger artifact-control protocols. Second, large-scale paired EEG--face datasets remain scarce, and the domain gap from controlled recordings to in-the-wild settings may affect generation quality under variations in identity, pose, background, or EEG acquisition conditions. Third, MindAU currently focuses on single-AU image editing, while real expressions involve multi-AU combinations and temporal dynamics. Finally, current metrics do not fully capture fine-grained AU intensity, localized deformation accuracy, or alignment with EEG-implied expression strength, motivating more fine-grained metrics and human perceptual studies.


\section{Conclusion}

In this paper, we introduced \textbf{MindAU}, a unified framework for EEG-conditioned facial action-unit editing. Unlike prior brain-guided generation methods that mainly reconstruct externally perceived stimuli, MindAU targets fine-grained facial expression editing from EEG signals while preserving the identity of a reference face. To address this challenging setting, we proposed an AU-aware EEG Encoder for noise-robust and AU-discriminative representation learning, a Dual-Stream Manifold Alignment module that aligns EEG features with both AU-level text semantics and identity-reduced visual displacement trajectories, and an EEG-conditioned facial editing backbone with EEG-aware M-RoPE, landmark-guided reference masking, and AU-aware region supervision. We further introduced \textbf{E-CAFE}, a curated benchmark with paired EEG--face editing samples and standardized protocols for self-referenced and cross-identity evaluation. Extensive experiments show that MindAU outperforms EEG-to-image generation and text-guided pipeline baselines, demonstrating the potential of EEG-conditioned facial editing for future assistive expression technologies.

\bibliographystyle{unsrt}
\bibliography{Reference}\label{reference}

\newpage
\appendix
\section{Landmark-Guided Progressive Masking Strategy}

\subsection{Reference-Image Masking}
In Stage 3, masking is applied only to the reference image, while the target image remains unmasked. This design aims to suppress direct appearance leakage from expression-relevant facial regions in the reference image, thereby encouraging the model to rely more heavily on EEG conditioning when reconstructing the target expression. During training, we adopt a linear masking-ratio schedule: at the beginning, 70\% of the reference images are masked over key facial regions and 30\% are left unchanged; the masking ratio is then linearly annealed until 3,000 training steps, after which 30\% of the reference images are masked and 70\% remain unaltered.

We use the standard 68-point facial landmark representation to define six local facial subregions that are closely related to facial expressions. Specifically, the landmark indices are grouped as follows: left eye, right eye, left eyebrow, right eyebrow, nose, and mouth. These regions are used to construct expression-relevant masks based on the activated Action Units (AUs).

\begin{table}[h]
\centering
\caption{Definition of local facial regions based on 68 facial landmarks. Landmark indices follow the standard 68-point annotation protocol.}
\label{tab:facial_regions}
\begin{tabular}{ll}
\toprule
Facial region & Landmark indices \\
\midrule
Left eye & 36--41 \\
Right eye & 42--47 \\
Left eyebrow & 17--21 \\
Right eyebrow & 22--26 \\
Nose & 27--35 \\
Mouth & 48--67 \\
\bottomrule
\end{tabular}
\end{table}

To determine which facial regions should be masked, we map each activated AU to its corresponding local facial region. For AUs related to eyebrow motion, including AU1, AU2, and AU4, both eyebrow regions are selected. AU5 is associated with the eye regions, AU9 with the nose region, and mouth-related AUs, including AU12, AU15, AU17, AU25, and AU27, are mapped to the mouth region.

\subsection{Geometric Construction of Expression-Relevant Masks}
\label{sec:mask_geometry}

Given the AU-selected facial regions, we construct geometric masks from the corresponding 68-point facial landmarks. The goal is to mask not only the sparse landmark locations, but also the surrounding expression-related appearance cues, such as wrinkles, local contours, and skin deformations.

For the eyes, eyebrows, and mouth, we use expanded elliptical masks. Let 
\(\mathcal{P}_r=\{(x_i,y_i)\}_{i=1}^{N_r}\) denote the landmark set of region \(r\). We first compute the axis-aligned bounding box of the landmarks and define its center and half sizes as
\begin{equation}
c_x = \frac{x_{\min}+x_{\max}}{2}, \quad
c_y = \frac{y_{\min}+y_{\max}}{2},
\end{equation}
\begin{equation}
h_w = \frac{x_{\max}-x_{\min}}{2}, \quad
h_h = \frac{y_{\max}-y_{\min}}{2}.
\end{equation}
The initial ellipse axes are obtained by region-specific scaling factors:
\begin{equation}
a_0 = s_x h_w, \quad b_0 = s_y h_h,
\end{equation}
where \(a_0\) and \(b_0\) denote the horizontal and vertical semi-axes, respectively. We further allow a small region-specific center offset:
\begin{equation}
c_x \leftarrow c_x + o_x a_0, \quad
c_y \leftarrow c_y + o_y b_0.
\end{equation}

To ensure that all landmarks are covered, we first enlarge the axes according to the maximum landmark deviation:
\begin{equation}
a = \max \left(a_0, \; m_{\mathrm{cover}} \max_i |x_i-c_x| \right),
\end{equation}
\begin{equation}
b = \max \left(b_0, \; m_{\mathrm{cover}} \max_i |y_i-c_y| \right).
\end{equation}
We then verify landmark inclusion under the normalized ellipse equation:
\begin{equation}
\rho = \max_i \left[
\left(\frac{x_i-c_x}{a}\right)^2 +
\left(\frac{y_i-c_y}{b}\right)^2
\right].
\end{equation}
If \(\rho > 1\), the ellipse axes are further enlarged by a factor of \(\sqrt{\rho}\), ensuring that all landmarks lie inside the ellipse:
\begin{equation}
a \leftarrow \sqrt{\rho}a, \quad b \leftarrow \sqrt{\rho}b.
\end{equation}
Finally, we apply an additional post-expansion factor \(m_{\mathrm{post}}\):
\begin{equation}
a \leftarrow m_{\mathrm{post}}a, \quad b \leftarrow m_{\mathrm{post}}b.
\end{equation}
The resulting elliptical mask for region \(r\) is defined as
\begin{equation}
\mathcal{M}_r =
\left\{
(u,v) \; \middle| \;
\left(\frac{u-c_x}{a}\right)^2 +
\left(\frac{v-c_y}{b}\right)^2
\leq 1
\right\}.
\end{equation}

The region-specific parameters used in our implementation are listed in Table~\ref{tab:ellipse_mask_params}.

\begin{table}[h]
\centering
\caption{Parameters for expanded elliptical masks. The same parameters are used for the left and right instances of eyes and eyebrows.}
\label{tab:ellipse_mask_params}
\begin{tabular}{lcccccc}
\toprule
Region & \(s_x\) & \(s_y\) & \(o_x\) & \(o_y\) & \(m_{\mathrm{cover}}\) & \(m_{\mathrm{post}}\) \\
\midrule
Eyes  & 1.55 & 1.85 & 0.00 &  0.00 & 1.10 & 1.18 \\
Eyebrows & 1.75 & 2.80 & 0.00 & -0.18 & 1.12 & 1.18 \\
Mouth & 1.90 & 2.20 & 0.00 &  0.12 & 1.12 & 1.18 \\
\bottomrule
\end{tabular}
\end{table}

For the nose region, we use a convex-hull mask rather than an ellipse:
\begin{equation}
\mathcal{M}_{\mathrm{nose}} = \mathrm{ConvHull}(\mathcal{P}_{\mathrm{nose}}).
\end{equation}
This design is sufficient because the nose region is relatively compact and exhibits less non-rigid deformation compared with the eyes, eyebrows, and mouth.

Given the set of facial regions selected by activated AUs, the final expression-relevant mask is obtained as the union of all corresponding regional masks:
\begin{equation}
\mathcal{M} = \bigcup_{r \in \mathcal{R}_{\mathrm{AU}}} \mathcal{M}_r,
\end{equation}
where \(\mathcal{R}_{\mathrm{AU}}\) denotes the set of regions determined by the AU-to-region mapping.

\subsection{Ablation Study}

\begin{figure}[ht]
  \centering\includegraphics[width=0.75\linewidth]{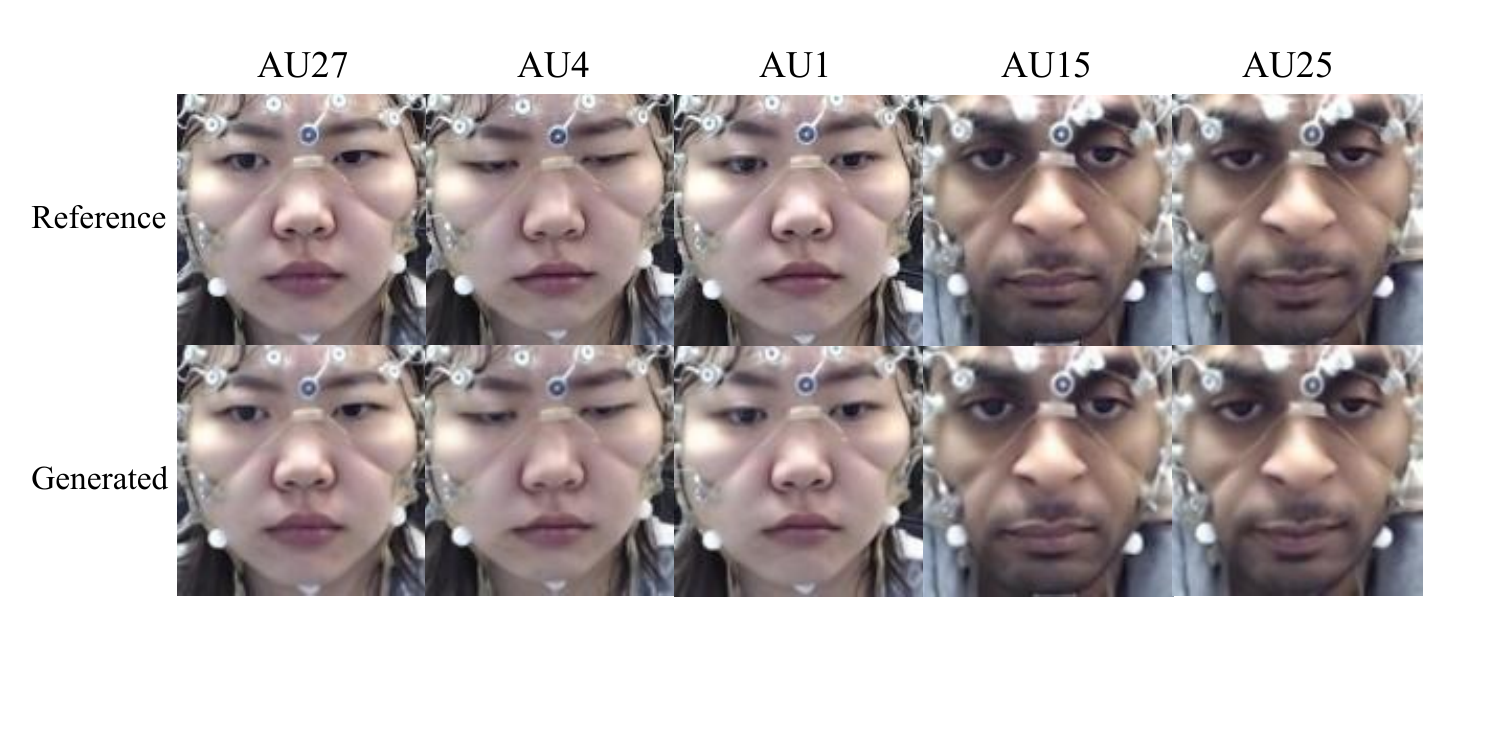}
  \vspace{-30pt}
  \caption{Qualitative comparison of the masking ablation under the self-reference setting.}
  \vspace{-5pt}
  \label{fig:abmask}
\end{figure}

We conducted an ablation study on the use of the masking strategy. The results without the masking strategy are presented in Table \ref{tab:ablation_mask}. It can be observed that while the CSIM is very high and the FID is remarkably low, the accuracy is significantly poor. As shown in Figure \ref{fig:abmask}, we found that without the masking strategy, the model suffers from severe "shortcut learning," where it simply replicates the reference image.

\begin{table}[t]
\centering
\caption{Ablation study on expression-relevant masking under the self-reference setting.}
\label{tab:ablation_mask}
\begin{tabular}{lcccc}
\toprule
Method & CSIM(\%) $\uparrow$ & FID $\downarrow$ & AU ACC(\%) $\uparrow$ & CLIP Score $\uparrow$ \\
\midrule
w/o Masking (Shortcut) & \textbf{90.45} & \textbf{35.12} & 5.02 & 15.49 \\
Ours (Full Model) & 75.33 & 43.35 & \textbf{27.71} & 17.61 \\
\bottomrule
\end{tabular}
\end{table}

\section{Implementation Details}

All experiments are conducted on 8*NVIDIA RTX A6000 GPUs with 48GB memory. Stage 1 takes approximately 2 GPU-hours, Stage 2 takes approximately 8*36 GPU-hours, and Stage 3 takes approximately 8*48 GPU-hours. 

\textbf{Stage 1.} We employ a Transformer-based asymmetric Masked Autoencoder (MAE)\cite{mae} with a 6-layer encoder and a 4-layer decoder. The model is pre-trained on EEG signals with an input dimensionality of $128 \times 48$. To facilitate robust temporal representation learning, we apply a 50\% masking ratio and optimize using AdamW for 80 epochs with a batch size of 32.

\textbf{Stage 2.} We align the EEG encoder with the Qwen2.5-VL \cite{qwen2.5vl} model via a multi-task learning framework. Optimization is performed using AdamW for 5,000 steps with a global batch size of 64. To mitigate memory bottlenecks, we pre-extract and cache visual features for both reference and ground-truth images using the frozen Qwen2.5-VL vision encoder \cite{qwen2.5vl}. This strategy bypasses the need to load the heavy vision backbone during alignment, significantly accelerating training.

\textbf{Stage 3.} We initialize the model using pre-trained parameters from LongCat-Image \cite{longcat-image}.  The model is trained for 5,000 steps at $512 \times 512$ resolution with a per-GPU batch size of 8. 

We further construct two natural pipeline baselines with LongCat-Image~\cite{longcat-image} and FireRed-Image-Edit~\cite{firered}, decomposing EEG-conditioned editing into EEG-to-AU recognition followed by text-guided reference-based editing. In these pipelines, the downstream text-guided editors are kept frozen and used in a plug-and-play setting with the same reference image and fixed AU text prompts. These baselines evaluate a practical off-the-shelf alternative to EEG-conditioned editing, rather than a task-specific retraining of large text-guided editing models. For a fair comparison on the EEG decoding component, the EEG-to-AU classifier shares the same Stage-1 EEG encoder and AU classification head as MindAU, achieving 32.89\% standalone AU accuracy on the held-out test split.

\section{E-CAFE Construction Details}
\begin{figure*}[ht]
  \centering\includegraphics[width=0.85\linewidth]{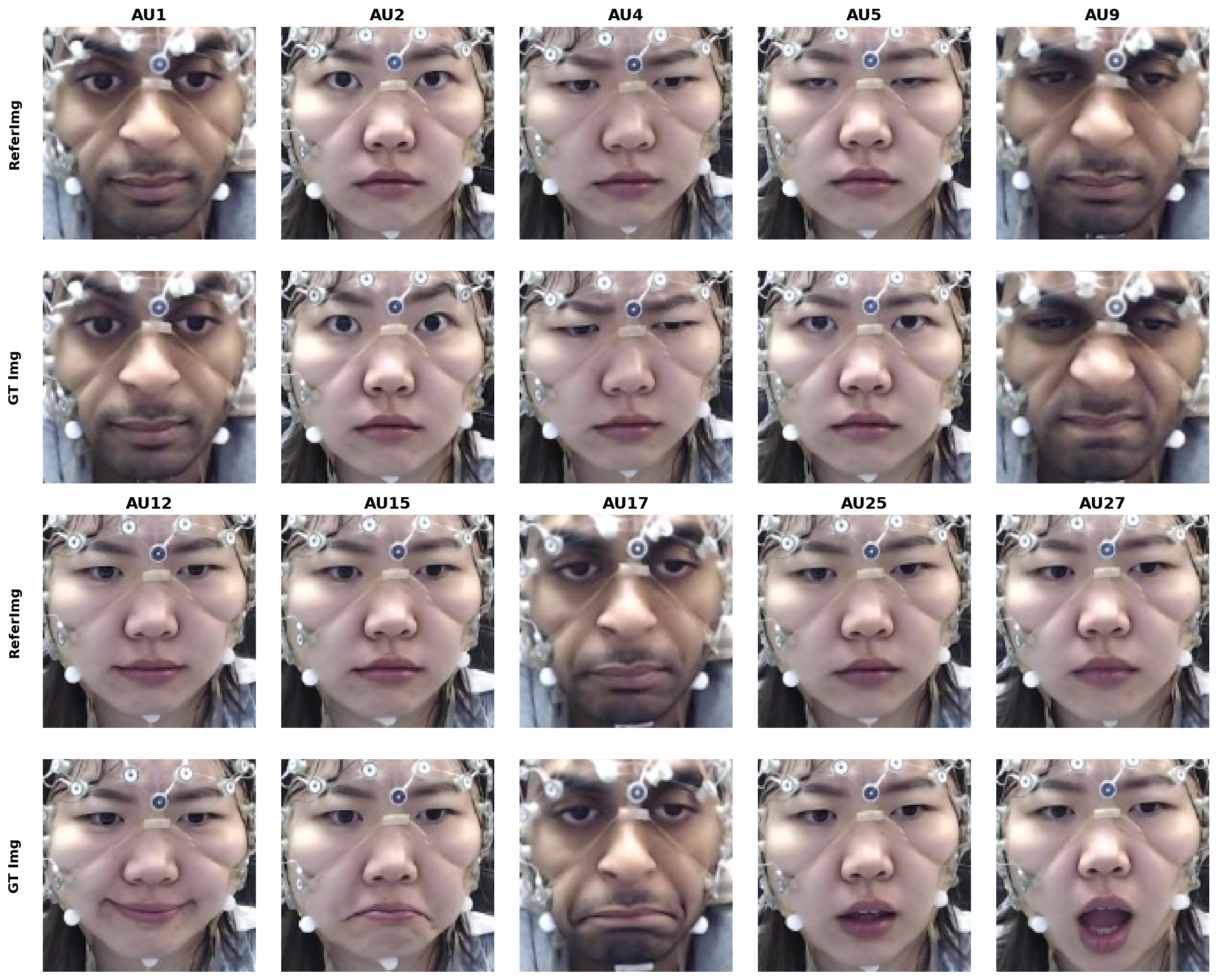}
  
  \caption{Visualization of samples from the Self-Referenced Editing test subset.}
  \vspace{-10pt}
  \label{fig:dataset}
\end{figure*}

In this section, we present a detailed visualization of E-CAFE. Fig.~\ref{fig:dataset} illustrates a subset of the test dataset for Self-Referenced Editing. To improve the reliability of EEG--image supervision, we employ a two-stage filtering process: landmark-based automatic pre-filtering followed by manual verification. We first extract 68-point facial landmarks for each candidate target frame using an offline face-alignment detector and discard samples with missing frames, failed face detection, or invalid landmarks. We then compute simple AU-specific geometric scores from normalized landmarks, such as brow--eye distance, eye-opening height, nose--upper-lip deformation, mouth width, and inner-mouth opening, to remove weak or ambiguous AU activations.

\begin{figure*}[ht]
  \centering\includegraphics[width=0.85\linewidth]{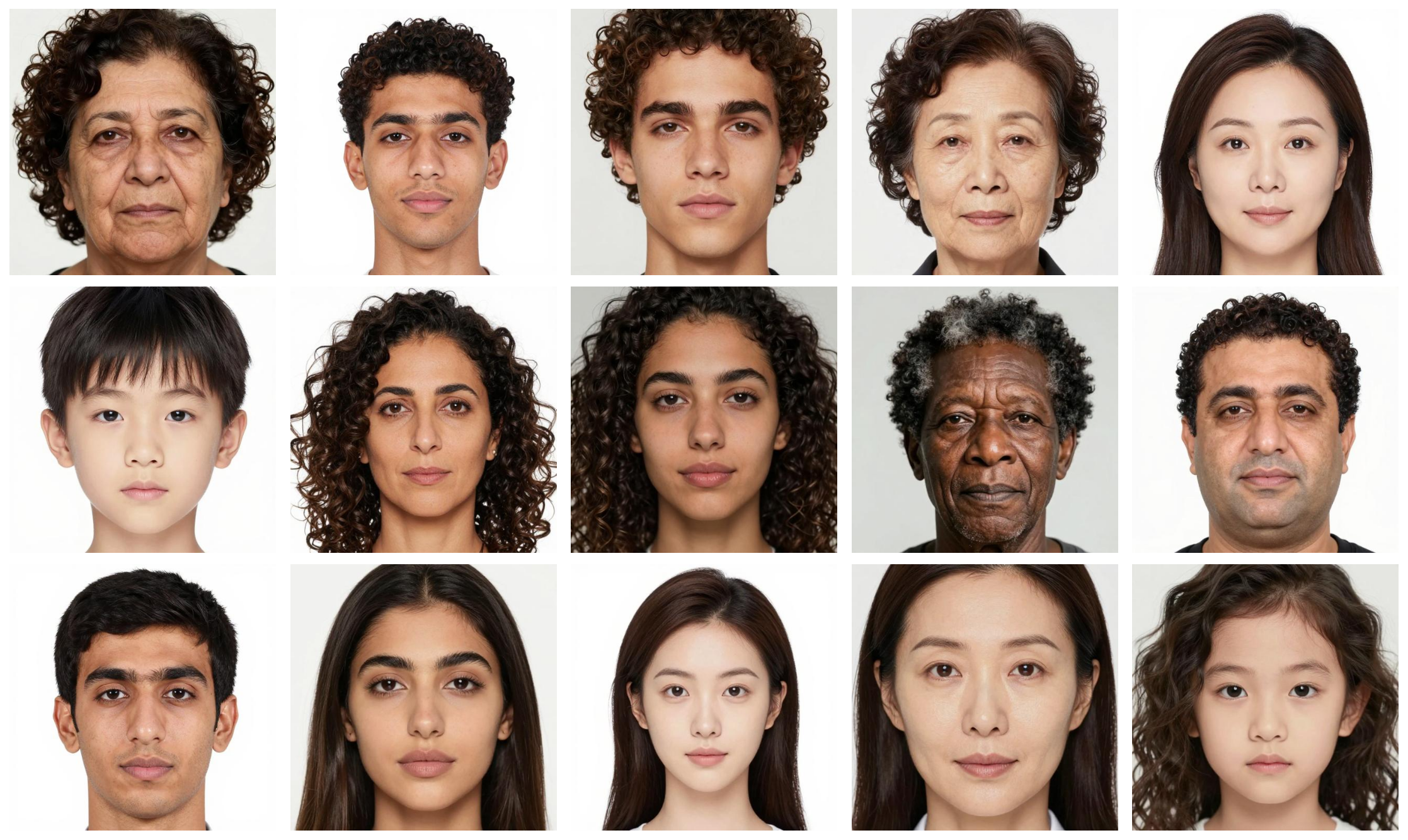}
  
  \caption{Visualization of reference faces from a subset of the Cross-Referenced Editing test dataset.}
  
  \label{fig:face}
\end{figure*}

The remaining samples are manually verified to remove residual landmark failures, ambiguous expressions, and reference frames with expression leakage. This filtering procedure improves the quality of EEG--image supervision by retaining visually reliable and AU-consistent pairs for both EEG-to-expression alignment and final image editing. Fig.~\ref{fig:face} displays a subset of face images generated by Z-Image~\cite{zimage} used in Cross-Referenced Editing.

\section{AU Text Descriptions}

Table~\ref{tab:au_text_descriptions} lists the AU-specific text descriptions used in this work. These descriptions serve as textual semantic targets for the Semantic Projection Stream in Stage 2, where they are encoded by the frozen Qwen2.5-VL text encoder for EEG--text alignment. The same descriptions are also used as fixed AU prompts for the text-guided pipeline baselines to ensure consistent semantic conditioning across methods.

\begin{table}[t]
\centering
\caption{Action unit (AU) labels and their corresponding text descriptions used in this work.}
\label{tab:au_text_descriptions}
\begin{tabular}{ll}
\toprule
\textbf{AU} & \textbf{Text Description} \\
\midrule
AU1  & inner brows raised \\
AU2  & outer brows raised \\
AU4  & brows lowered and drawn together \\
AU5  & upper eyelids raised \\
AU9  & nose wrinkled \\
AU12 & lip corners pulled up \\
AU15 & lip corners pulled down \\
AU17 & chin raised \\
AU25 & lips parted \\
AU27 & mouth wide open \\
\bottomrule
\end{tabular}
\end{table}

\section{More Visualization}
\begin{figure*}[ht]
  \centering\includegraphics[width=0.85\linewidth]{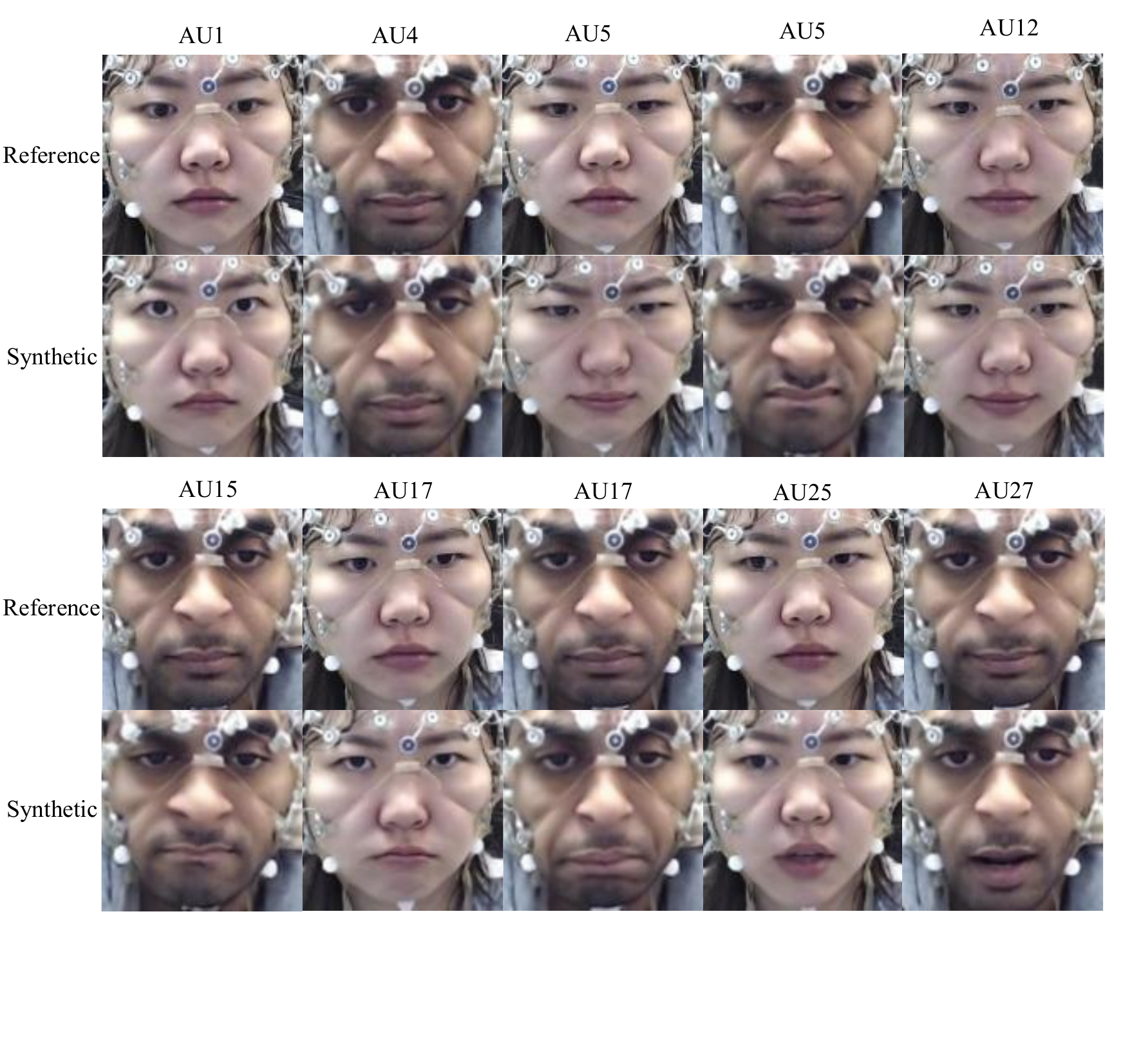}
  \vspace{-40pt}
  \caption{Qualitative results of Self-Referenced Editing on the test dataset.}
  \vspace{-30pt}
  \label{fig:more}
\end{figure*}

\begin{figure*}[ht]
  \centering\includegraphics[width=0.85\linewidth]{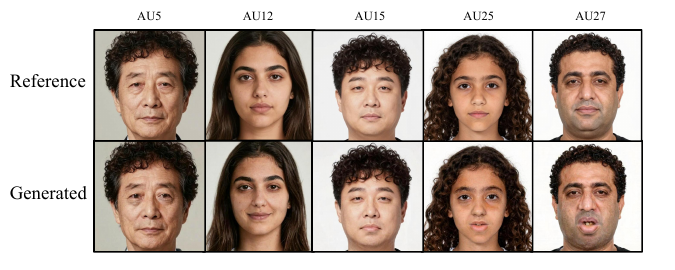}
  \vspace{-10pt}
  \caption{Qualitative results of Cross-Referenced Editing on the test dataset.}
  
  \label{fig:more2}
\end{figure*}

Fig. \ref{fig:more} and Fig. \ref{fig:more2} present additional visualization results.

\section{Dual-Stream Manifold Alignment}

\begin{figure*}[t] 
    \centering
    \begin{subfigure}[b]{0.48\linewidth}
        \centering
        \includegraphics[width=0.9\linewidth]{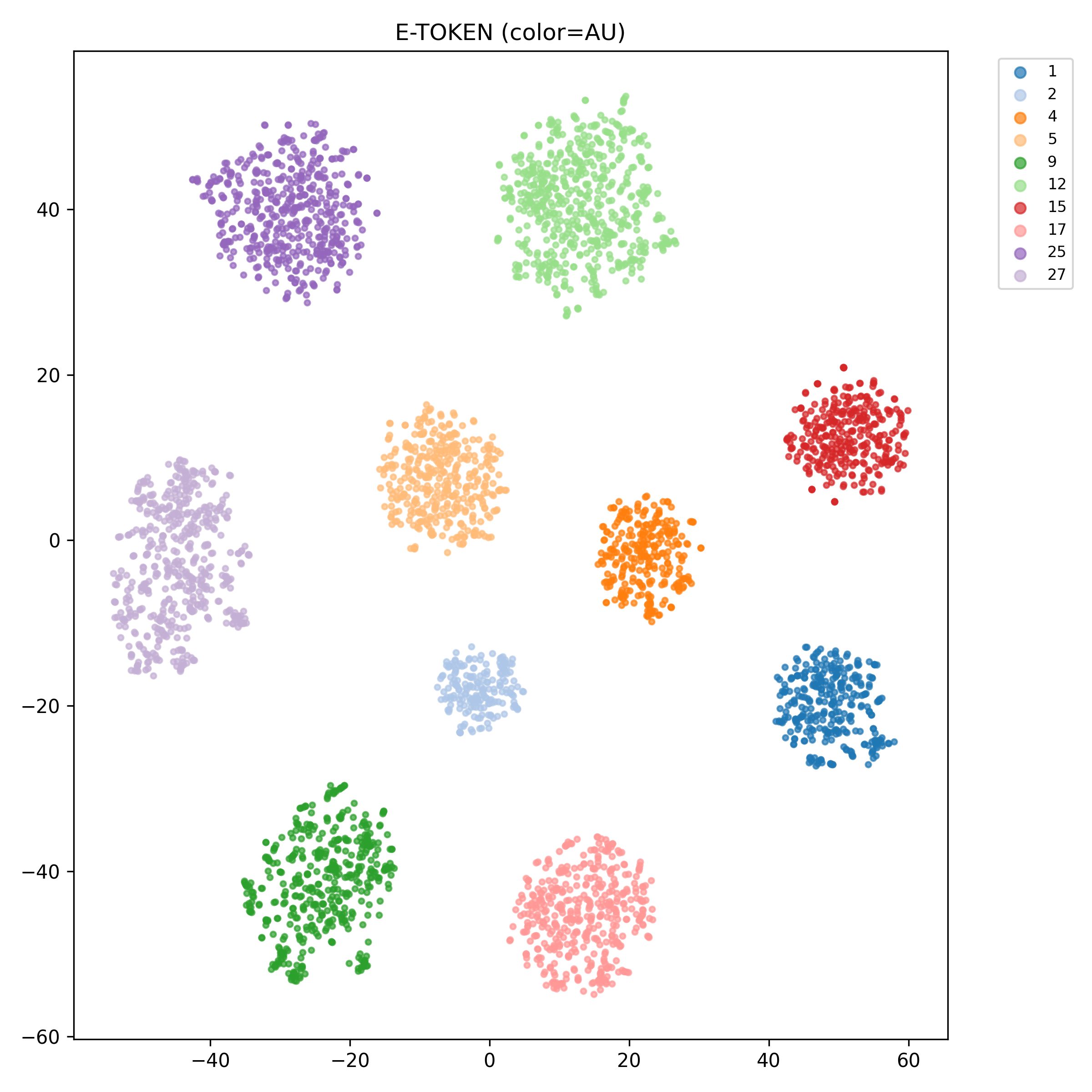}
        \label{fig:eeg_au}
    \end{subfigure}
    \hfill 
    \begin{subfigure}[b]{0.48\linewidth}
        \centering
        \includegraphics[width=0.9\linewidth]{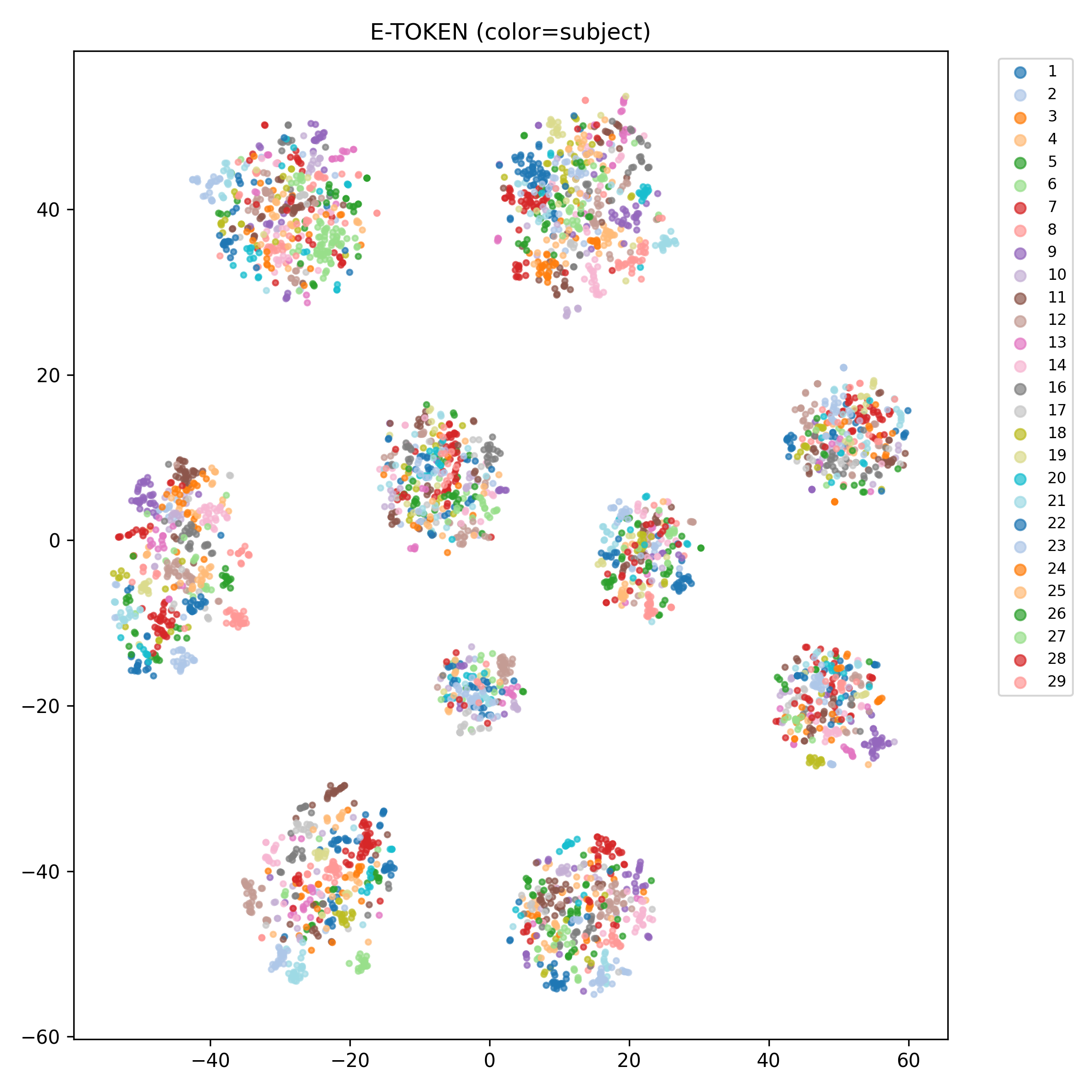}
        \label{fig:eeg_sub}
    \end{subfigure}
    \vspace{-5pt}
    \caption{Visualization of E-token feature clustering extracted by the EEG adapter on the training set. Left: labeled by AU; Right: labeled by Subject.}
    \vspace{-5pt}
    \label{fig:eeg_combined}
\end{figure*}

As shown in Fig. \ref{fig:eeg_combined}, we visualize the features of EEG tokens extracted from the train set. The results indicate that our EEG Adapter effectively discriminates between distinct AU features while eliminating interference from different subjects. To investigate the impact of utilizing generated pre-training data in the second stage, we conducted an ablation study, as shown in Table \ref{tab:eegpretain}. The results demonstrate that our full model outperforms the variant without pre-training (w/o pretrain) across key metrics, achieving a higher CSIM, a lower FID, and an improved AU ACC.

\begin{table}[t]

\centering

\caption{Ablation study on the effectiveness of generated pre-training data in the second stage.}

\label{tab:eegpretain}

\begin{tabular}{lcccc}

\toprule

Method & CSIM(\%) $\uparrow$ & FID $\downarrow$ & AU ACC(\%) $\uparrow$ & CLIP Score $\uparrow$ \\

\midrule

Ours (w/o pretrain) & 74.29 & 45.22 & 27.10 & 17.98 \\

Ours (Full Model) & 75.33 & 43.35 & 27.71 & 17.61 \\

\bottomrule

\end{tabular}

\end{table}

\section{Failure Case Analysis}

\begin{figure*}[ht]
  \centering
  \includegraphics[width=0.85\linewidth]{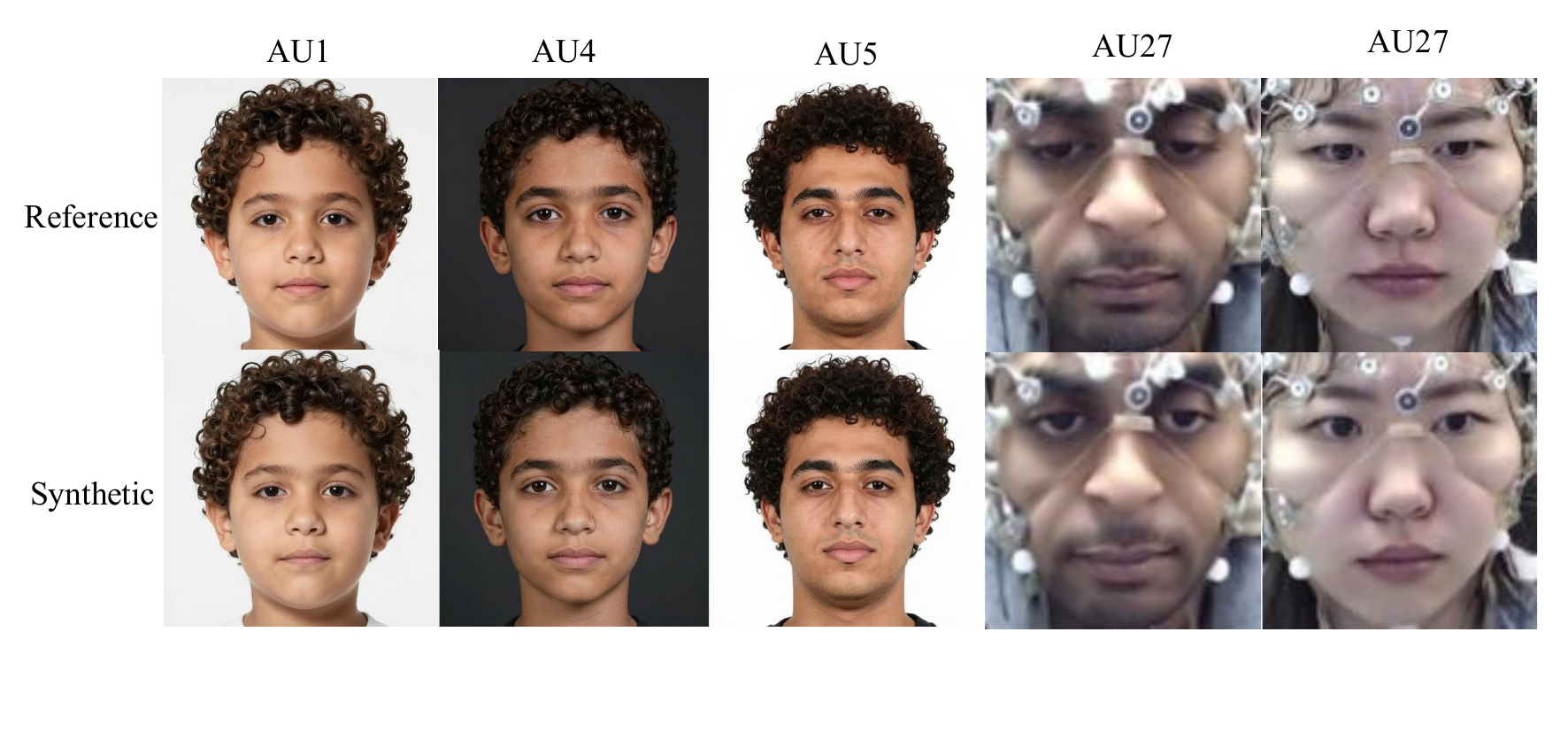}
  \vspace{-20pt}
  \caption{Representative failure cases of MindAU. Some outputs exhibit incorrect AU generation or insufficient expression change, where the generated image remains overly close to the reference image.}
  \vspace{-10pt}
  \label{fig:fail}
\end{figure*}

Figure~\ref{fig:fail} shows representative failure cases of MindAU. Although the proposed landmark-guided reference masking strategy reduces reference-dominant shortcut learning, the model can still occasionally under-utilize EEG conditions when the target expression is subtle or the reference and target images are visually similar. Besides incorrect AU generation, we observe a small number of cases where the output nearly copies the reference image, producing insufficient expression change. This suggests that fully preventing reference shortcut behavior remains challenging under limited paired EEG--face supervision. 

\section{Inference Details}

\subsection{Inference Pipeline}

At inference time, we follow the same conditioning structure used during Stage-3 training. 
Given a reference face image and an input EEG segment, we first preprocess the reference image into two branches: 
a $512\times512$ branch for the VAE encoder and a $256\times256$ branch for the Qwen2.5-VL encoder. 
The text condition uses the same fixed prompt as training, namely \textit{``a face showing facial action unit''}. 
This fixed text prompt is encoded together with the reference image by the Qwen2.5-VL encoder, and we retain the prompt-token hidden states corresponding to the editable text span.

The EEG signal is processed by the Stage-1 EEG Encoder and Stage-2 EEG adapter. The adapter maps the EEG input into a sequence of 32 E-tokens, each with hidden dimension 3584. 
These E-tokens are appended to the text prompt embeddings, yielding the final multimodal conditioning sequence. 
We keep distinct modality identifiers for text and EEG tokens, so that the diffusion transformer can distinguish linguistic and EEG conditions while attending to both within a unified context space.

The reference image is encoded by the VAE into latent space, while the target latent is initialized from Gaussian noise. We use the same FlowMatch Euler scheduler as in training-time validation, with 30 denoising steps for train--test consistency. 
After denoising, the predicted latent is decoded by the VAE and resized back to the original reference-image resolution for saving.

\subsection{EEG-Free Inference}

To isolate the contribution of EEG conditioning, we further introduce an EEG-free inference setting. 
In this setting, the overall inference pipeline remains unchanged, including the reference image, fixed text prompt, LoRA checkpoint, and sampling configuration. 
The only modification is applied to the EEG branch: instead of feeding the EEG-derived tokens, we replace the appended E-token block with an all-zero token matrix of the same shape:
\[
\mathbf{E}_{\text{free}}=\mathbf{0}\in\mathbb{R}^{N_e\times d},
\]
where $N_e=32$ denotes the number of EEG tokens and $d=3584$ is the token embedding dimension.

As shown in Table~\ref{tab:eegfree}, removing EEG conditioning leads to a substantial performance degradation across all evaluation metrics. 
Specifically, the EEG-free setting obtains only 25.12\% CSIM and 7.22\% AU ACC, while the full model achieves 75.33\% and 27.71\%, respectively. 
Meanwhile, the FID increases sharply from 43.35 to 136.2, indicating a clear decline in image quality and distribution fidelity.

\begin{table}[t]
\centering
\caption{Comparison between EEG-free inference and EEG-conditioned model under self-reference protocol.}
\label{tab:eegfree}
\begin{tabular}{lcccc}
\toprule
Method & CSIM(\%) $\uparrow$ & FID $\downarrow$ & AU ACC(\%) $\uparrow$ & CLIP Score $\uparrow$ \\
\midrule
EEG-free & 25.12 & 136.2 & 7.22 & 14.43 \\
Ours (Full Model) & 75.33 & 43.35 & 27.71 & 17.61 \\
\bottomrule
\end{tabular}
\end{table}





\section{Asset Licenses and Terms of Use}

Table~\ref{tab:asset_licenses} summarizes the existing datasets, models, and software assets used in this work. We use these assets only for research purposes and respect their original licenses or terms of use. Unless explicitly allowed by the corresponding license or data-use agreement, we do not redistribute third-party raw data or model weights. For assets whose public license information is not available in a standard open-source format, we follow the original access agreement and direct users to obtain the asset from the original provider.

\begin{table}[t]
\centering
\small
\caption{Existing assets used in this work and their licenses or terms of use.}
\label{tab:asset_licenses}
\begin{tabular}{p{4.0cm} p{9.5cm}}
\toprule
\textbf{Asset} & \textbf{License or terms of use} \\
\midrule
BU-EEG 
& Original BU-EEG data-use terms / access agreement. \\

Qwen2.5-VL-7B-Instruct
& Apache-2.0 for the 7B checkpoint. If a different Qwen2.5-VL variant is used, we follow the license attached to that specific checkpoint. \\

LongCat-Image-Edit
& Apache-2.0. \\

FireRed-Image-Edit-1.0
& Apache-2.0 for code and weights. \\

Z-Image
& Apache-2.0. \\

Gemini / Nano Banana 2
& Google Gemini API Additional Terms of Service. \\

Face Alignment Network
& BSD-3-Clause. \\

OpenGraphAU
& Apache-2.0. \\

ArcFace / InsightFace
& MIT for code; released pretrained models may be restricted to non-commercial research use depending on the checkpoint. \\
\bottomrule
\end{tabular}
\end{table}



\end{document}